\pdfoutput=1

\documentclass[11pt]{article}

\usepackage{styles/eacl2023}
\usepackage{times}
\usepackage{latexsym}
\usepackage{tikz}
\usepackage[T1]{fontenc}
\usepackage[utf8]{inputenc}
\usepackage{microtype}

\usepackage{inconsolata}

\RequirePackage[inline,shortlabels]{enumitem}
\setlist{nosep}
\RequirePackage[disable,color=yellow!10,linecolor=red,textsize=scriptsize]{todonotes}
\RequirePackage{amsmath,amssymb,bm}
\RequirePackage{adjustbox}

\makeatletter
\g@addto@macro{\normalsize}{%
\setlength{\abovedisplayskip}{2pt minus1pt}%
\setlength{\abovedisplayshortskip}{2pt minus1pt}%
\setlength{\belowdisplayskip}{2pt minus1pt}%
\setlength{\belowdisplayshortskip}{2pt minus1pt}}

\textfloatsep \intextsep
\floatsep \intextsep
\makeatother

\usepackage{times}
\usepackage{latexsym}
\usepackage{url}
\usepackage{graphicx}
\usepackage{multirow}
\usepackage{amsmath}
\usepackage{amssymb}
\usepackage{balance}
\usepackage{varwidth}
\usepackage{subfig}
\usepackage{multirow}

\newcommand{\mdl}{TwiRGCN}
\newcommand{\cronkgqa}{CronKGQA}
\newcommand{\exaqt}{EXAQT}

\newcommand{\data}{\textit{TimeQuestions}}

\newcommand{\mtq}{m_{tq}}
\newcommand{\qbert}{q_B}
\newcommand{\tq}{q_t}

\newcommand{\refeqn}[1]{Eqn.\,\eqref{#1}}
\newcommand{\reffig}[1]{Figure\,\ref{#1}}
\newcommand{\reftbl}[1]{Table\,\ref{#1}}
\newcommand{\refsec}[1]{Section\,\ref{#1}}
\newcommand{\refappendix}[1]{Appendix\,\ref{#1}}

\def\ztitle{\mdl: Temporally Weighted Graph Convolution for Question Answering over Temporal Knowledge Graphs}

\title{\ztitle}

\author{Aditya Sharma \\
  IISc, Bangalore \\
  \texttt{adityasharma@iisc.ac.in} \\\And
  Apoorv Saxena \\
  IISc, Bangalore \\
  \texttt{apoorvsaxena@iisc.ac.in} \\\And
  Chitrank Gupta \\
  IIT Bombay \\
  \texttt{chigupta2011@gmail.com} \\\AND
  Mehran Kazemi \\
  Google Research, Montreal \\
  \texttt{mehrankazemi@google.com} \\\And
  Partha Talukdar \\
  Google Research, India \\
  \texttt{partha@google.com} \\\And
  Soumen Chakrabarti \\
  IIT Bombay \\
  \texttt{soumen@cse.iitb.ac.in}
  }

\begin{document}
\maketitle

\begin{abstract}

Recent years have witnessed interest in Temporal Question Answering over Knowledge Graphs (TKGQA), resulting in the development of multiple methods. However, these are highly engineered, thereby limiting their generalizability, and they do not automatically discover relevant parts of the KG during multi-hop reasoning. Relational graph convolutional networks (RGCN) provide an opportunity to address both of these challenges -- we explore this direction in the paper. Specifically, we propose a novel, intuitive and interpretable scheme to modulate the messages passed through a KG edge during convolution based on the relevance of its associated period to the question. We also introduce a gating device to predict if the answer to a complex temporal question is likely to be a KG entity or time and use this prediction to guide our scoring mechanism. We evaluate the resulting system, which we call TwiRGCN, on a recent challenging dataset for multi-hop complex temporal QA called \data{}. We show that TwiRGCN significantly outperforms state-of-the-art models on this dataset across diverse question types. Interestingly, TwiRGCN improves accuracy by 9--10 percentage points for the most difficult ordinal and implicit question types.

\end{abstract}

\section{Introduction}



Question answering (QA) is a key problem in natural language processing and a long-lasting milestone for artificial intelligence. 
A large class of approaches for QA makes use of knowledge graphs (KG), which are multi-relational graphs representing facts (KGQA). 
Temporal KGs (TKG) represent facts that are only valid for specific periods of time as \textit{(subject, relation, object, time range)}, for example, \textit{(Franklin\ D\ Roosevelt, position\ held, President of USA, $[1933, 1945]$)}.
The problem of answering questions that require temporal reasoning over TKGs (TKGQA) is a special case of KGQA that specifically focuses on the following challenge: temporal questions constrain answers through temporal notions, e.g., ``\textit{who was the first president of US during WW2?}'' 
Developing systems for temporal QA is of immense practical importance for many applications.
It is considered a more challenging problem than KGQA \citep{bhutani2019learning, saxena-etal-2020-improving}, where questions are typically about persistent, non-temporal facts (e.g., place of birth), with only a small portion of the questions requiring any temporal reasoning \cite{zhen-etal-2018-tempquestions}.

\begin{figure}[thb]
    \centering
    \includegraphics[width=.95\hsize]{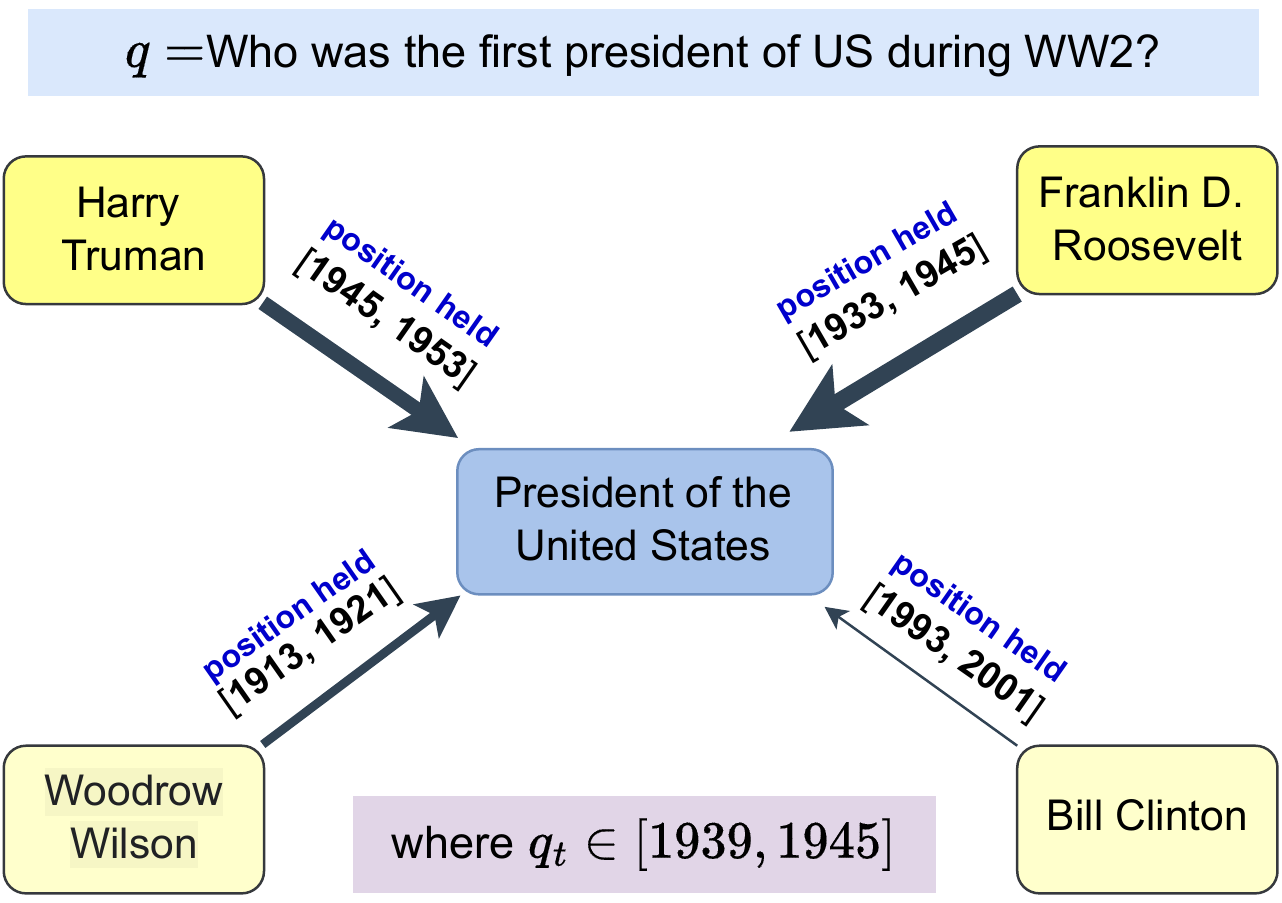}
    \caption{\label{fig:msg_pass} \small An illustrative example of how our temporal gating described in \refsec{subsec:twirgcn} modulated the incoming graph convolution messages for one node depending on the time period of interest for the question. The thickness of an edge here is proportional to the value of the temporal edge weight $m_{tq}^{(e)}$ for that edge. In this example, the entities \emph{Franklin D.\  Roosevelt} and \emph{Harry Truman}, who were presidents during WW2 $[1939, 1945]$ get the top two highest weights, while \emph{Woodrow Wilson}, who was president during WW1 $[1914, 1918]$ gets a smaller edge weight. In contrast, \emph{Bill Clinton}, whose time period is unrelated to the question, gets a much lower edge weight. Thus, contributing very little to the convolution update of the 'President of the US' node.} 
\end{figure}

Even though a variety of models have been proposed for the TKGQA recently, they suffer from the following problems: 1)~they are either highly engineered toward the task \citep{exaqt, ref:science_dir_subgtr} or 2)~they do not incorporate graph structure information using Graph Neural Networks (GNN) \citep{ ref:tempoqr, shang-etal-2022-tsqa, ref:cron_kgqa}. 
We explore the following hypotheses in this paper: 1)~a simple GNN-based solution could generalize better and offer higher performance than highly engineered GNN-based, and TKG embedding-based models; 2)~a multi-layer GNN model could do multi-hop reasoning across its layers; 3)~not all edges (temporal facts) are equally important for answering temporal questions (see \reffig{fig:msg_pass}), so GNN solutions could benefit from temporally weighted edge convolutions.


Following the aforementioned hypotheses, we develop a novel but architecturally simple TKGQA system that we call ``Temporally weighted Relational Graph Convolutional Network'' (\mdl{}).  It is based on the Relational Graph Convolutional Network (RGCN) proposed by \citet{ref:rgcn}. \mdl{} introduces a question-dependent edge weighting scheme that modulates convolutional messages passing through a temporal fact edge based on how relevant the time period of that edge is for answering a particular question.
In RGCN, convolution messages from all TKG edges are weighted equally. But all edges are not equally important for answering temporal questions. For example, in \reffig{fig:msg_pass}, to answer the question ``\textit{Who was the first president of the US during WW2?}'' the edge with Bill Clinton has little relevance for answering the question. But, regular RGCN would still weigh all edges equally. We address this shortcoming through our proposed modulation. We impose soft temporal constraints on the messages passed during convolution, amplifying messages through edges close to the time period relevant for answering the question while diminishing messages from irrelevant edges. This leads to better, more efficient learning as we are not confusing our model with unnecessary information, as evidenced by our significantly improved performance without the need for any heavy engineering.
We explore two different strategies for our convolutional edge weighting, which show complementary strengths.
Our experiments establish that \mdl\ significantly outperforms already strong baselines on \data. 
Our contributions are:
\begin{itemize}[leftmargin=*]

%
\item We propose \mdl{}, a simple and general TKGQA system that computes question-dependent edge weights to modulate RGCN messages, depending on the temporal relevance of the edge to the question.
\item We explore two novel and intuitive schemes for imposing soft temporal constraints on the messages passed during convolution, amplifying messages through edges close to the time relevant for answering the question while diminishing messages from irrelevant edges. We also propose an answer-gating mechanism based on the likelihood that the answer is an entity or time.
\item Through extensive experiments on a challenging real-world dataset, we find that \mdl{} substantially outperforms prior art in  overall accuracy, and by 9--10\% on the implicit and ordinal type questions --- categories that require significant temporal reasoning.
\item We augment \data{} with a TKG and release both code and data at \href{https://github.com/adi-sharma/TwiRGCN}{https://github.com/adi-sharma/TwiRGCN}.
\end{itemize}

\section{Related Work}
\label{sec:Rel}

Most KGQA systems have focused on answering questions from simple (i.e., 1-hop fact-based questions) \cite{berant-etal-2013-semantic} to multi-hop complex questions requiring multi-fact reasoning \cite{ref:pullnet, saxena-etal-2020-improving}. However, only a small fraction of these questions require any temporal reasoning \cite{zhen-etal-2018-tempquestions}. Recent efforts have tried to overcome this gap by proposing models as well as datasets to explicitly focus on temporal reasoning.  We review these below.

\noindent \textbf{Temporal KGQA methods:} One line of work uses temporal constraints along with hand-crafted rules to find the answer \citep{bao-etal-2016-constraint, luo-etal-2018-knowledge, zhen-etal-2018-tequila}.  A recent class of models has leveraged advances in TKG embedding methods for answering questions on Temporal KGs.   CronKGQA \citep{ref:cron_kgqa} does this by posing a question as a TKG completion problem and finds the answer using the TComplex \citep{Lacroix2020Tensor} score function and BERT \citep{ref:bert} question embedding to complete the fact. TempoQR \citep{ref:tempoqr} uses additional temporal supervision to enrich TKG embeddings, followed by a transformer-based decoder \citep{ref:transformer}. TSQA \citep{shang-etal-2022-tsqa} on the other hand estimate the time in the question and uses it to enrich TKG embeddings for finding the answer. SubGTR \citep{ref:science_dir_subgtr} infers question-relevant temporal constraints using TKG embeddings and applies them as filters to score entities in the question subgraph.
Although we, too, use pre-trained TKG embeddings to initialize our generalized RGCN, we use the GNN framework to take advantage of the structural information in the KG in ways that they do not. Recent work \citep{teru2020inductive} shows that GNN-based models can encode any logical rule corresponding to a path in the knowledge graph. We refer to this as structural information that shallow embedding-based models cannot access.

\noindent \textbf{RGCN based QA systems:}
Graph neural networks are increasingly being used in QA systems not specifically meant for temporal reasoning.  GraftNet \cite{ref:graft_net} uses personalized PageRank to collect a query-relevant subgraph from a global KG, then an RGCN to predict the answer from the relevant subgraph.  PullNet \cite{ref:pullnet} loops over and expands GraftNet's subgraph to do multi-hop reasoning.  EXAQT \citep{exaqt} is the system closest to ours: it addresses TKGQA and also uses an RGCN.  
The RGCN for answer prediction which works on the question subgraph is very similar to that in GraftNet. EXAQT augments it with dictionary matching, heavy engineering, and additional category information. In contrast, \mdl{} uses a straightforward temporally weighted graph convolution followed by answer gating, as described in \refsec{sec:method}, while still achieving superior performance (see \refsec{subsec:quant}).
More details in \refsec{subsec:baselines}.

\section{Preliminaries}
\label{sec:background}


\subsection{Temporal Knowledge Graphs (TKG)}
\label{subsec:def_temp_kg}

\noindent \textbf{KG:} Multi-relational graphs with entities (eg: Barack Obama, USA) as nodes and relations $r$ between entities $\{s,o\}$ (e.g., president of) represented as typed edges between nodes.  Each edge of this graph, together with endpoint nodes, represents a fact triple $\{s,r,o\}$, e.g., $\{\text{Barack Obama}, \text{president of}, \text{USA}\}$.

\noindent \textbf{TKG:} Numerous facts in the world are not perpetually true and are only valid for a certain time period. A TKG represents such  a fact as a quadruple of the form $\{s,r,o,[t_{st},t_{et}]\}$, where $t_{st}$ is the start time and $t_{et}$ is the end time of validity of the fact, e.g., $\{\text{Barack Obama}, \text{president of}, \text{USA}, [2009, 2017]\}$.

\subsection{Question Answering on TKGs}
\label{subsec:def_tkgqa}

Given a question $q$ specified in natural language form and a TKG $\mathcal{G}$, TKGQA is the task of finding the answer to $q$ based on the information that is available (or can be derived) from $\mathcal{G}$.  
A subgraph of $\mathcal{G}$ is a subset of its nodes with induced edges.
In this paper, we assume each question is already associated with a subgraph $\mathcal{G}_q $ relevant to the question. We define $\mathcal{G}_q = (\mathcal{V}_q, \mathcal{R}_q, \mathcal{T}_q, \mathcal{E}_q)$ as the subgraph of $\mathcal{G}$ associated with a question $q\in\mathcal{Q}$, where $\mathcal{Q}$ represents the set of all questions.
Each edge $e \in \mathcal{E}_q$ represents a fact $\{v_i, r, v_j, [t_{st}, t_{et}]\}$, where $v_i, v_j \in \mathcal{V}_q$ are entity nodes, $r\in\mathcal{R}_q$ is the relation between them and $t_{st}, t_{et} \in \mathcal{T}_q$ are the start and end times for which the fact is valid.


\subsection{Relational Graph Convolutional Networks}
\label{subsec:rgcn}


Given a KG, each node $v_i$ is initialized to a suitable embedding $h_{v_i}^{(0)}$ at layer~0.Thereafter, \citet{ref:rgcn} propose to update node embeddings $h_{v_i}^{(l+1)}$, at layer $(l+1)$, as follows:
\begin{align}
h_{v_i}^{(l+1)} \hspace{-.2em} &= \sigma\hspace{-.2em}
\left( \sum_{r \in \mathcal{R}}\sum_{j \in \mathcal{N}^r_i}
\hspace{-.3em}
\frac{W_r^{(l)} h_{v_j}^{(l)}}{|\mathcal{N}_i^r|} + W_0^{(l)}h_{v_i}^{(l)}
\!\right)
\label{eq:rgcn}
\end{align}
where $\mathcal{N}_i^r$ is the set of neighbors of node $v_i$ that are connected via relation edges of type~$r$, $\mathcal{R}$ is the set of relations,  $W_r^{(l)}$ are weight matrices associated with each relation type $r$ and layer~$l$. They are initialized using a basis decomposition method.


\section{Proposed Method: \mdl{}}
\label{sec:method}

In this section, we develop and describe \mdl{} (``Temporally Weighted Relational Graph Convolutional Network''), our model for TKGQA. 
\begin{figure*}[thb]
    \centering
    \includegraphics[width=.95\hsize]{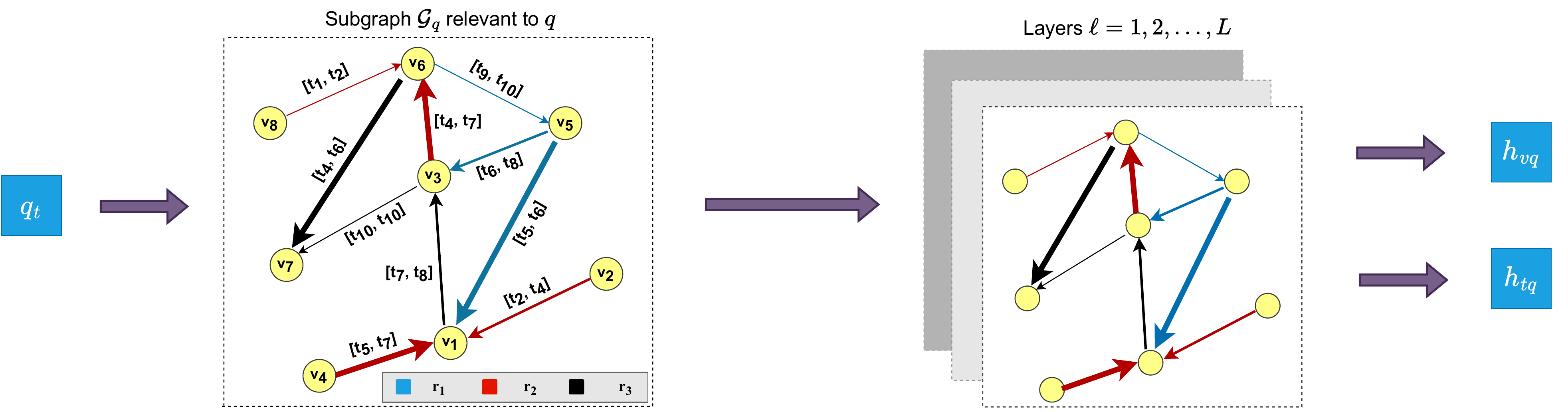}
    \caption{\label{fig:twirgcn_pool} \small \textbf{Left:} Shows temporally weighted convolutional message passing described in \refsec{subsec:twirgcn} happening across a subgraph $G_q$ for one layer. For the same, we get question-dependent temporal edge weights $\mtq^{(e)}$ using \emph{question time}, $\tq$  (described in \ref{subsec:m_types}). \textbf{Right:} As discussed in \refsec{subsec:twirgcn}, embeddings are propagated in the subgraph $G_q$ for a fixed number of layers $(L)$ and hidden units of the final layer are pooled to get entity prediction, $h_{vq}$. We get time prediction, $h_{tq}$, by pooling the updated embeddings for all unique times in $G_q$.}
\vspace{-.4cm}
\end{figure*}


\subsection{Embedding for questions and KG facts}
\label{subsec:emb}

\noindent \textbf{Question embedding:} 
We pass the question text through a pre-trained encoder-only language model (LM) to obtain a question embedding. In particular, we prepend a [CLS] token to the input question and feed it into BERT \cite{devlin-etal-2019-bert}, and then use its output-layer [CLS] embedding as the question embedding $\qbert{}$.  We enable LM fine-tuning during training.

\noindent \textbf{TKG preprocessing for RGCN initialization:} 
We initialize entity and time embeddings using pre-trained TComplEx \cite{Lacroix2020Tensor} embeddings.\footnote{TComplEx is known to provide high-quality embeddings, but other TKG embedding methods such as TimePlex \citep{TimePlex} can also be used.} To obtain these for the \data{} dataset \cite{exaqt}, we first construct a `background KG' $\mathcal{G} =  \bigcup_{q \in \mathcal{Q}} \mathcal{G}_q$ which is the union of all question subgraphs $\mathcal{G}_q$ in the train dataset. 
As in most temporal KGQA works, we discretize time to a suitable granularity (in our dataset, a year).\footnote{\mdl{} can be extended to TKGQA datasets that do not provide subgraphs through recently proposed subgraph selection methods \citep{ref:science_dir_subgtr, shang-etal-2022-tsqa}.}
The graph on which \mdl{} is run represents every entity as a node~$v_i$ and time as edge attribute ~$t_j$.  Their initial (layer-0) RGCN embeddings $h_{v_i}^{(0)}$ and $h_{t_j}$, are set to the entity and time embeddings obtained from TComplEx, respectively.
We refer to $h_{tj}$ as $h_{st}^{(e)}$ and $h_{et}^{(e)}$ depending on $t_j$ appearing as start or end time for edge $e$, respectively.
When $e=(i,r,j)$, we will use superscript $(i,r,j)$ in place of~$(e)$.

\begin{figure}[t]
    \centering
    \includegraphics[width=.95\hsize]{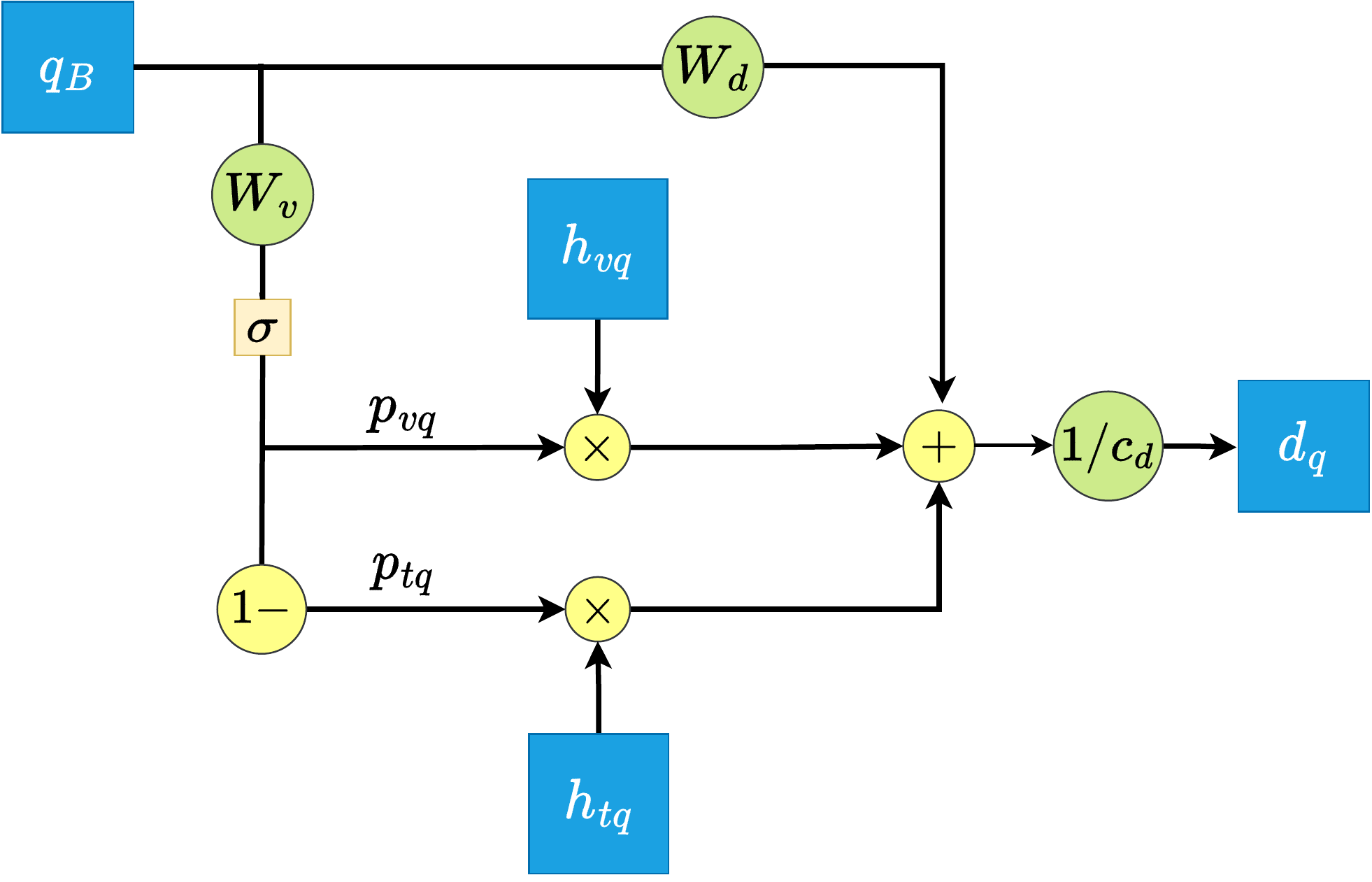}
    \caption{\label{fig:ans_gate} \small Predicting the answer based on gating entity prediction ($h_{vq}$) and time prediction ($h_{tq}$) of a subgraph $G_q$ based on the likelihood that the answer is either an entity ($p_{vq}$) or time ($p_{tq}$) given question $q$, respectively. Details in \refsec{subsec:gating}.} 
\end{figure}

\subsection{Temporally modulated edge weights}
\label{subsec:twirgcn}

Having available the question subgraph, and the initial entity and time embeddings, our system applies a temporally weighted graph convolution on the local subgraph to enable answering questions that require complex temporal reasoning over a KG.
To achieve this, 
we introduce 
a question-dependent temporal edge weight $\mtq^{(i,r,j)} \in [-1,1]$ for modulating the convolutional message passed through edge $e$ valid from time $t_{st}^{(i,r,j)}$ to $t_{et}^{(i,r,j)}$ connecting node $v_i$ to $v_j$ via relation $r$, $\{v_i,\ r,\ v_j,\ [t_{st}^{(i,r,j)}, t_{et}^{(i,r,j)}]\}$ which assigns a weight to that edge depending on how relevant the time period of $e$ is for answering question $q$. Then, motivated by \refeqn{eq:rgcn}, we update the hidden state for a node $v_i$ in the temporal KG at layer $(l{+}1)$ as:
\begin{multline}
h_{v_i}^{(l+1)}= \sigma \bigg( W_0^{(l)}h_{v_i}^{(l)} + \\[-1ex]
\sum_{r \in \mathcal{R}}
\sum_{j \in \mathcal{N}^r_i}
{\setlength{\fboxsep}{1pt}
\colorbox{yellow!30}{$\mtq^{(i,r,j)}$}} \frac{W_r^{(l)} h_{v_j}^{(l)}}{|\mathcal{N}_i^r|} \bigg).
\label{eq:twi_rgcn}
\end{multline}
See \reffig{fig:msg_pass} for an example update for one node.



As shown in \reffig{fig:twirgcn_pool}, after passing messages across a subgraph $G_q$ over $L$ such layers, we pool the hidden states from the final layer of all nodes in $G_q$ to get $h_{vq}$, the entity prediction. Similarly, we pool the updated embeddings for all unique times in $G_q$ to get $h_{tq}$, the time prediction. We describe in \refsec{subsec:gating} how we use $h_{vq}$ and $h_{tq}$ to get the final predicted answer from our model. We use mean pooling in this work, but any other pooling operation can also be used.


\subsection{Edge weighting formulations}
\label{subsec:m_types}

We explore two different formulations for computing $\mtq^{(i,r,j)}$, namely \emph{average} and \emph{interval}, and discuss the motivations behind the two approaches. 
In \refsec{sec:eval}, we empirically show that the inductive bias inherent in each of the two approaches makes them excel at different types of temporal reasoning while giving similar performance overall. We also provide an intuitive explanation of how the edge weighting formulations of the two approaches explain the difference between their empirical results.
We first project the question embedding $\qbert$, using a learned projection matrix $W_{tq}$, to find the \emph{question time} embedding $\tq = W_{tq} \qbert$. In the following, $\mtq^{(i,r,j)}$ = $\mtq^{(e)}$ is the weight for edge $e$.

\subsubsection{\mdl~(average)}

In this variant, we calculate the edge modulation $\mtq$ as the cosine similarity between the question time embedding, and the average of the embeddings for the start and the end time of an edge:
\begin{align}
\mtq^{(e)} &= \cos\left(\frac{h_{st}^{(e)} + h_{et}^{(e)}}{2},\, \tq \right).
\label{eq:twirgcn_average}
\end{align}
This formulation gives a high weight to an edge if the question time falls close to the middle of the time interval for an edge. For example, if the edge times are [2008, 2012] and the question time is 2010, the edge is weighted highly.

\subsubsection{\mdl~(interval)} 
In this variant, $\mtq^{(e)}$ is defined as the mean of two cosine similarities:
\begin{enumerate*}[(1)]
\item the cosine similarity between the start time of the edge and the learned question time embedding, and
\item the cosine similarity between the end time of the edge and the learned question time embedding.
\end{enumerate*}
Formally,
\begin{align}
\mtq^{(e)} &= \frac{\cos(h_{st}^{(e)}, \tq)
{+}\cos(h_{et}^{(e)}, \tq)}{2}.
\label{eq:twirgcn_interval}
\end{align}
This formulation weighs an edge highly if question time $t_q$ lies within the time interval of the edge.

\noindent \textbf{Generality beyond temporal reasoning} While we developed \mdl{} for temporal reasoning, the edge weighting is more general and could extend to the case where $\tq{}$ is a "goal" embedding for any goal-directed task.


\subsection{Answer type gating}
\label{subsec:gating}

Question answering over TKGs may involve questions whose answer is an entity (e.g., \emph{Who was ...?}) or whose answer is a time (e.g., \emph{When did ...?}). We hypothesize that it should be possible to predict whether the answer to a question is an entity or a time based on the text of the question; making such a prediction helps filter out (or down-weight) a portion of the nodes in that graph that are less likely to be the answer. Toward this hypothesis, we introduce a gating mechanism that learns the likelihood that the answer is an entity $p_{vq}$ or a time $p_{tq}$ given the question: 
\begin{align}
p_{vq} &= 1 - p_{tq} = \sigma (w_{v} \qbert),
\label{eq:gating}
\end{align}
where $w_v$ transforms $\qbert$ to a scalar and $\sigma$ is the sigmoid function that ensures $0 \leq p_{vq} \leq 1$.
As shown in \reffig{fig:ans_gate}, we then compute a prediction embedding $d_q$ for question $q$ as a gated sum of the entity prediction and time prediction (see \refsec{subsec:twirgcn} and \reffig{fig:twirgcn_pool}) added to the question embedding:
\begin{align}
d_q = \frac{1}{c_d} [p_{vq} h_{vq} + p_{tq} h_{tq} + W_d \qbert],
\label{eq:prediction}
\end{align}
where $c_d$ is a constant hyperparameter and $W_d$ is the weight for transforming $\qbert$ to the dimension of the entity and time embeddings. Having the prediction embedding $d_q$, we rank candidate answers (entities and times from the global TKG) based on their similarity to~$d_q$.




\noindent \textbf{Training} We score all possible answer entities and times as a cosine distance with the prediction embedding ($d_q$), scaled using a constant hyperparameter. We take a softmax over all these scores and train using the cross-entropy loss.

\section{Evaluation}
\label{sec:eval}



\subsection{Dataset}
\label{subsec:dataset}

Earlier works on TKGQA use the automatically generated CronQuestions dataset \citep{ref:cron_kgqa}. A recent analysis, however, shows that this dataset comes with several limitations that stem from its automatic construction method \citep{ref:science_dir_subgtr}. Specifically, there are spurious correlations in the dataset that can be exploited by different models to achieve high accuracy (e.g., \citet{ref:tempoqr} report more than $90\%$ accuracy overall and $99\%$ in some categories on this dataset). Therefore, we base our experiments on a recent more challenging dataset, namely \data\ \citep{exaqt}, where the aforementioned models perform poorly (as seen in \reftbl{tbl:results_table}). 


\begin{table}
\centering \begingroup \small
\begin{tabular}{p{0.15\linewidth} | p{0.75\linewidth}}
\hline
\textbf{Category}                                           & \multicolumn{1}{c}{\textbf{Question}}                                              \\ \hline
\multirow{2}{*}{\textbf{Explicit}}                      & \textit{Which team won the 2010 F1 world championship?}                            \\
                                                        & \textit{What honour did Agatha Christie win in 1971?}                              \\ \hline
\multirow{2}{*}{\textbf{Implicit}}                      & \textit{Who did Kevin Garnett play for before Celtics?}                            \\
                                                        & \textit{Where was Leonardo Da Vinci when he died?}                                 \\ \hline
\multicolumn{1}{l|}{\multirow{2}{*}{\textbf{Temporal}}} & \textit{What years did the team with fight song Steelers polka win the Superbowl?} \\
\multicolumn{1}{l|}{}                                   & \textit{What year did Sam Elliott and Kathryn Ross marry?}                         \\ \hline
\multirow{2}{*}{\textbf{Ordinal}} & \textit{What was the first satellite to maintain orbit around the earth in space?} \\
\multicolumn{1}{l|}{}                                   & \textit{What is the third book of the twilight series?}                            \\ \hline
\end{tabular} 
\endgroup
\caption{\label{tbl:data_table}
Examples of questions from each category in \data\ dataset, discussed in \refsec{subsec:dataset}.
}
\end{table}


\noindent \textbf{\data} has $13.5k$ manually curated questions divided into the train, valid, and test splits containing $7k$, $3.2k$, and $3.2k$ questions, respectively. The questions fall under four types: `Explicit,' `Implicit,' `Temporal,' and `Ordinal,' based on the type of temporal reasoning required to answer the questions. We show some examples of questions from each of these categories in \reftbl{tbl:data_table}.
We augment this dataset with question-specific subgraphs generated from WikiData in the final step of the answer graph construction pipeline proposed by \citet{exaqt}. We preprocess all the obtained facts to the \textit{(subject, relation, object, [start time, end time])} format, and restrict all times to years, a format used by most contemporary TKGs. 
We create a "background KG'' described in \refsec{subsec:emb} as a union of all subgraphs in the train set. This background KG contains $240k$ facts, $118k$ entities, and $883$ relations. 
We include this augmented \data{} dataset and associated code at \href{https://github.com/adi-sharma/TwiRGCN}{https://github.com/adi-sharma/TwiRGCN}.


\begin{table}
\centering \begingroup \small
\begin{tabular}{p{0.18\linewidth} | p{0.72\linewidth}}
\hline
\textbf{Dataset}               & \multicolumn{1}{c}{\textbf{Question}}                                                   \\ \hline
\textbf{ComQA}                 & \textit{Who played Dumbledore in the 5th harry potter film?}                            \\ \hline
\textbf{Complex-Web-Questions} & \textit{What is the name of the club the subject of "golden shoes" played for in 2010?} \\ \hline
\textbf{Graph-Questions}       & \textit{What sports were in both the 1912 summer Olympics and the 2008 Olympic games?}  \\ \hline
\textbf{LC-QuAD 2.0}           & \textit{What is the start time for Heidi Klum has spouse as Seal?}                      \\ \hline
\textbf{Free917}               & \textit{What is the price of a 2012 jeep wrangler sport?}                               \\ \hline
\end{tabular}
\endgroup
\caption{\label{tbl:component_table} \small
Examples of questions requiring temporal reasoning from KGQA datasets (see \refsec{subsec:dataset}).
}
\end{table}


\noindent \textbf{Temporal subsets of KGQA datasets:} \data{} is a compilation of temporal questions from different KGQA datasets. We show the results in \reftbl{tbl:results_component} on a subset of 5 such datasets included in the test set of \data{} namely, ComQA \citep{ref:comqa}, ComplexWebQuestions \citep{ref:ComplexWebQuestions}, GraphQuestions \citep{ref:GraphQuestions}, LC-QuAD 2.0 \citep{ref:lcquad2}, and Free917 \citep{ref:free917}. \reftbl{tbl:component_table} shows representative examples from these 5 datasets.


\begin{table*}
\centering
\adjustbox{max width=.9\hsize}{ \small
\begin{tabular}{l|lllll}
\hline
\multicolumn{1}{l|}{} & \multicolumn{1}{l}{Overall} & \multicolumn{1}{l}{Explicit} & \multicolumn{1}{l}{Implicit} & \multicolumn{1}{l}{Temporal} & Ordinal \\ \hline
PullNet \citep{ref:pullnet}                                                                              & 0.105                        & 0.022                         & 0.081                         & 0.234                             & 0.029   \\
Uniqorn \citep{ref:uniqorn}                                                                              & 0.331                        & 0.318                         & 0.316                         & 0.392                             & 0.202   \\
GRAFT-Net \citep{ref:graft_net}                                                                               & 0.452                        & 0.445                         & 0.428                         & 0.515                             & 0.322   \\
\cronkgqa \ \citep{ref:cron_kgqa}                                                                                & 0.462                        & 0.466                         & 0.445                         & 0.511                             & 0.369   \\
TempoQR \citep{ref:tempoqr}                                                                                & 0.416                        & 0.465                         & 0.36                          & 0.4                               & 0.349   \\ 
\exaqt \ \citep{exaqt}                                                                                 & 0.572                        & 0.568                         & 0.512                         & 0.642                             & 0.42    \\ \hline
  \textbf{\mdl{}} \textit{(average)}                                                                           &  \textbf{0.605}                        &  \textbf{0.602}                         &  0.586                         &  0.641                             &  \textbf{0.518}   \\
 \textbf{\mdl{}} \textit{(interval)}                                                                           &  \textbf{0.603}                        &  0.599                         &  \textbf{0.603}                         &  \textbf{0.646}                             &  0.494   \\
 \hline
\end{tabular} }
\caption{\label{tbl:results_table} \small
Comparison of Hits@1 for different Temporal KGQA methods on \data\ dataset (\refsec{subsec:quant}). Interestingly, \mdl{} improves accuracy over SOTA by $3.3\%$ overall and by $9$-$10\%$ for the most difficult ordinal \& implicit question types.
}
\end{table*}


\subsection{Baseline methods}
\label{subsec:baselines}

We compare TwiRGCN against a spectrum of existing methods, including EXAQT, other TKGQA methods, and non-temporal KGQA methods.

\noindent \textbf{Non-temporal KGQA methods:} We include Unicorn \citep{ref:uniqorn}, which uses Group Steiner Trees for answering questions. We test on two RGCN-based approaches for KGQA, namely, GRAFT-Net \citep{ref:graft_net}, which attends over relations of neighborhood edges based on the question, and PullNet \citep{ref:pullnet}, which extends GRAFT-Net for multi-hop questions.

\noindent \textbf{TKGQA methods:} We also compare against TKGQA methods CronKGQA \citep{ref:cron_kgqa} and TempoQR \citep{ref:tempoqr} recently proposed for the CronQuestions dataset.  In contrast to \mdl, these do not leverage the powerful GNN framework. CronKGQA frames QA as a KG completion problem to complete the fact the question is interested in, using the TComplex score function and BERT question embedding. TempoQR, on the other hand, enriches pre-trained TKG embeddings with additional supervision from the dataset and uses a transformer \citep{ref:transformer} based decoder to predict the final answer.

\noindent \textbf{EXAQT:} \citet{exaqt} propose EXAQT, which is hitherto the best performer on \data.  It is also an RGCN-based TKGQA model that utilizes the GRAFT-Net framework. But in contrast to our model, EXAQT is heavily engineered. It utilizes the ground truth question category information from the dataset at train and test time, so it always knows whether the answer is temporal or belongs to another category. In contrast, our model learns the likelihood that the answer is an entity or time without any explicit supervision through our gating mechanism described in \refsec{subsec:gating}. EXAQT also uses explicit temporal signals from the question, extracted through a dictionary matching-based method using predefined temporal words such as `before', `after', `first', `last', `during', etc.  It then enriches its embeddings by utilizing the above in a multi-step end-to-end process.
In contrast, our models do not have access to any such information with only a straightforward temporally weighted graph convolution followed by answer gating, as described in \refsec{sec:method}.


\begin{table}
\centering \begingroup \small
\begin{tabular}{p{0.45\linewidth} | c | c}
\hline
\textbf{}            & \textbf{TwiRGCN} & \textbf{EXAQT} \\ \hline
ComQA                & \textbf{0.413}   & 0.292          \\
ComplexWebQuestions & \textbf{0.728}   & 0.515          \\
GraphQuestions       & \textbf{0.382}   & 0.323          \\ 
LC-QuAD 2.0          & 0.71             & \textbf{0.732} \\
Free917              & 0                & \textbf{0.17}  \\ \hline
\end{tabular}
\endgroup
\caption{\label{tbl:results_component} \small
Results for \exaqt{} and \mdl{} on temporal subsets of well-known KGQA datasets, as discussed in \refsec{subsec:quant}. \mdl{} beats \exaqt{} by a high margin up to 21\% on ComQA, ComplexWebQuestions, and GraphQuestions, which include questions requiring multi-hop reasoning.}
\end{table}


\subsection{Results}
\label{subsec:quant}

\noindent \textbf{\mdl{} achieves new state-of-the-art:} We compare the accuracy (Hits@1) for different Temporal KGQA models across all question categories found in \data{} in \reftbl{tbl:results_table}. From this table, we see that our models \mdl{} (average) and \mdl{} (interval) achieve significant improvements of up to $3.3\%$ overall absolute accuracy over the previous state-of-the-art model, \exaqt. Additionally, \mdl{} (average) gets a $9.8\%$ improvement over \exaqt \ in the ordinal category and \mdl{} (interval) improves over \exaqt \ by $9.1\%$ in the implicit category. The questions in both these categories require significant temporal reasoning to find the correct answer. Both models also show a marked improvement of up to $3.4\%$ in the explicit question category.

\noindent \textbf{\mdl{} (average) vs (interval):} Even though the two \mdl{} variants achieve comparable overall accuracy, they do so in different ways, showing complementary strengths.  \mdl~(average) achieves a $2.4\%$ improvement over \mdl~(interval) in the ordinal category, while \mdl~(interval) improves over \mdl~(average) for the implicit and temporal question types. We intuitively explain this behavior as a consequence of their edge weighting function~$\mtq$. In \mdl~(average), $\mtq$ (defined in \refeqn{eq:twirgcn_average}) is at its peak when the average time of edge is close to the question time, enabling it to reason between the temporal ordering of facts more effectively. Thus, helping it better answer ordinal questions of the type \emph{first}, \emph{fourth} or \emph{last occurrence}, etc. In contrast, $\mtq$ for \mdl~(interval), as defined in \refeqn{eq:twirgcn_interval}, helps answer questions that require temporal reasoning over specific times rather than just a temporal ordering of facts, which are mainly present in the implicit and temporal categories.

\noindent \textbf{Temporal subsets of KGQA datasets:} As mentioned earlier,  \data{} is a compilation of temporal questions from different KGQA datasets. To provide a finer-grained comparison, we compare \mdl{} to our most competitive baseline, \exaqt, on these subsets.
We show the results in Table~\ref{tbl:results_component}. 
TwiRGCN outperforms EXAQT by a high margin of up to 21\% in Hits@1 on the \emph{ComQA}, \emph{ComplexWebQuestions}, and \emph{GraphQuestions} datasets. These three datasets contain questions requiring complex multi-hop reasoning.
In contrast, we are competitive but perform slightly worse than EXAQT on \emph{LC-QuAD 2.0}, a templated dataset created from SPARQL queries, and \emph{Free917} that primarily consists of quantity-based questions. 
These results show our model’s generalizability and superior performance over our primary baseline on complex multi-hop questions. They also identify a failure mode for questions whose answers are quantities (explored in \refsec{subsec:quali}). 


\subsection{Analysis}
\label{subsec:quali}

Here we explore how our models behave qualitatively and look at example cases where they perform well and cases where they do not.




\begin{table}
\centering \begingroup \small
\adjustbox{max width=\hsize}{
\begin{tabular}{c|cc} \hline
& \textbf{With gating} & \textbf{W/o gating} \\ \hline
\textbf{\mdl~(average)}  & 0.605                & 0.597               \\ \hline
\textbf{\mdl~(interval)} & 0.603                & 0.597               \\ \hline
\end{tabular} }
\endgroup
\caption{\label{tbl:ablation} \small
Results of ablation study to see contributions of answer gating described in \refsec{subsec:gating} on overall Hits@1. We that it contributes about 0.7\% on average to the overall accuracy of our models
}
\end{table}


\begin{table}
\centering \begingroup \small
\begin{tabular}{cc}
\multicolumn{2}{c}{\textbf{Temporal Distance from $\tq$}} \\ \hline
\multicolumn{1}{c|}{\textbf{Median}}     & \textbf{5}    \\ \hline
\multicolumn{1}{c|}{=0}                & 18.3\%       \\ \hline
\multicolumn{1}{c|}{${\le}5$}       & 51.5\%        \\ \hline
\multicolumn{1}{c|}{${\le}20$}      & 74.8\%        \\ \hline
\end{tabular}
\endgroup
\caption{\label{tbl:cutie} \small
The median temporal distance from learned $\tq$ to extracted time is just 5 years, while we predict an exact match 18.3\% of the time (discussed in \refsec{subsec:quali}).
} 
\end{table}




\begin{table*}
\centering \begingroup \small
\adjustbox{max width=.8\hsize}{\small
\begin{tabular}{c|c|cc|cc}
\hline
\multirow{2}{*}{\textbf{\begin{tabular}[c]{@{}c@{}}Ground\\ Truth\end{tabular}}} & \multirow{2}{*}{\textbf{Prediction}} & \multicolumn{2}{c|}{\textbf{TwiRGCN (average)}} & \multicolumn{2}{c}{\textbf{TwiRGCN (interval)}}                   \\ \cline{3-6} 
& & \multicolumn{1}{c|}{with gating}  & w/o gating  & \multicolumn{1}{c|}{with gating} & \multicolumn{1}{l}{w/o gating} \\ \hline
Entity                                                                           & Time                                 & 3.18 \%                              & 7.28 \%       & 3.23 \%                             & 4.53 \%                           \\ \hline
Time                                                                             & Entity                               & 7.99 \%                              & 6.88 \%        & 7.3 \%                              & 7.82 \%                           \\ \hline
\end{tabular} }
\endgroup
\caption{\label{tbl:ent_time_gating} \small
Percentage of questions for which answer is an entity but our model incorrectly predicts time and vice versa. We analyze this in \refsec{subsec:quant} with and w/o answer gating 
to show that our proposed answer gating helps in reducing such mistakes.
}
\vspace{-.4cm}
\end{table*}


\noindent \textbf{Ablation for answer gating:} To understand the contribution of the proposed answer gating method described in \refsec{subsec:gating}, we perform an ablation study by removing answer gating from \refeqn{eq:prediction}.
By comparing columns of \reftbl{tbl:ablation} we can infer that our proposed answer gating contributes about 0.7\% on average to the overall accuracy of our models.

\noindent \textbf{How accurate is predicted question time?} We predict a question time embedding $\tq{}$ close to the time of interest for answering question q, as described in \refsec{subsec:m_types}. Here we analyze the effectiveness of this prediction by getting the time with embedding closest to $\tq$. 
We then use regex-based time extraction on questions and can extract time for 1199 questions in the test set. As seen in \reftbl{tbl:cutie}, out of a time range of 2916 years (including BC and AD years), our median distance from learned question time to extracted time is just 5 years while we predict an exact match 18.3\% of the time. Additionally, for 51.5\% of the questions the distance is ${\le}5$ years, while it is ${\le}20$ years for $\sim$75\% of the questions. 
Our simple-to-learn $\tq$ which is just a linear transform of $\qbert$ works reasonably well. 
Better $\tq$ would result in even more performance improvements. We leave that for future work.


\begin{table}
\centering
\begin{tabular}{cccc}
\hline
\multicolumn{1}{c|}{\textbf{}} & \multicolumn{1}{c|}{\textbf{Hits@1}} & \multicolumn{1}{c|}{\textbf{Hits@2}} & \textbf{Hits@3} \\ \hline
\multicolumn{4}{c}{EXPLICIT}                                                                                                   \\ \hline
\multicolumn{1}{c|}{EXAQT}     & \multicolumn{1}{c|}{0.568}           & \multicolumn{1}{c|}{0.602}           & 0.618           \\ \hline
\multicolumn{1}{c|}{TwiRGCN}   & \multicolumn{1}{c|}{\textbf{0.602}}  & \multicolumn{1}{c|}{\textbf{0.618}}  & \textbf{0.628}  \\ \hline
\multicolumn{4}{c}{IMPLICIT}                                                                                                   \\ \hline
\multicolumn{1}{c|}{EXAQT}     & \multicolumn{1}{c|}{0.512}           & \multicolumn{1}{c|}{0.575}           & 0.612           \\ \hline
\multicolumn{1}{c|}{TwiRGCN}   & \multicolumn{1}{c|}{\textbf{0.603}}  & \multicolumn{1}{c|}{\textbf{0.622}}  & \textbf{0.637}  \\ \hline
\multicolumn{4}{c}{ORDINAL}                                                                                                    \\ \hline
\multicolumn{1}{c|}{EXAQT}     & \multicolumn{1}{c|}{0.42}            & \multicolumn{1}{c|}{0.47}            & 0.49            \\ \hline
\multicolumn{1}{c|}{TwiRGCN}   & \multicolumn{1}{c|}{\textbf{0.518}}  & \multicolumn{1}{c|}{\textbf{0.542}}  & \textbf{0.553}  \\ \hline
\end{tabular}
\caption{\label{tbl:hitsk}
Effects of increasing k for Hits@k. As discussed in \refsec{subsec:quali}, \mdl{} significantly outperforms \exaqt{} across categories of questions even as k is increased.
}
\end{table}

\noindent \textbf{Dominant errors:} We do an error analysis over quantity-type questions, a challenging query class. Neither EXAQT nor TwiRGCN perform well on quantity-type questions. Out of a total of $224$ quantity questions, EXAQT gets $0.1\%$ accuracy while \mdl{} gets $0.05\%$. This is because current TKGQA models treat quantities such as “2.55” or “16,233” as independent entities, instead of scalar numeric values.  Additionally, from \refsec{subsec:quant}, we reconfirm that current TKGQA models fail on this bucket, so future work can direct special attention here. Examples: “\textit{What was Panama's fertility rate in 2006?}” A: 2.55; “\textit{What was the population of Bogota in 1775?}” A: 16,233.


\begin{table}
\centering
\begin{tabular}{c|ccc}
\hline
\textbf{Temporal (Hits@1)} & $\mathbf{\pm 0}$ & $\mathbf{\pm 1}$ & $\mathbf{\pm 3}$\\ \hline
EXAQT                   & 0.642           & 0.653                               & 0.667                     \\ \hline
TwiRGCN (average)       & 0.641                               & 0.649                               & 0.671 \\ \hline
TwiRGCN (interval)      & \textbf{0.646}     & \textbf{0.659}                         & \textbf{0.682}               \\ \hline
\end{tabular}
\caption{\label{tbl:time_period}
Effects of increasing the answer temporal window on model performance for Temporal type questions. As discussed in \refsec{subsec:quali}, \mdl{} (interval) gets even more accurate relative to \exaqt\ as we increase the temporal tolerance window.
}
\end{table}

\noindent \textbf{Reducing answer type mistakes:} In this study, we estimate \mdl's propensity to make answer type mistakes. We define these mistakes as questions where the answer was an entity, but our model predicted a time or vice versa.
From \reftbl{tbl:ent_time_gating} we see that our answer gating mechanism mentioned in \refeqn{eq:gating} helps reduce such mistakes. For \mdl~(average), gating cuts entity-to-time mistakes by more than half.


\noindent \textbf{Increasing $k$ for Hits@$k$:} We extend the analysis in \reftbl{tbl:results_table} increasing $k$ from 1 to 3 for Hits@$k$ on \data{}. From \reftbl{tbl:hitsk} we see that the performance of \mdl{} is robust to increasing~$k$. It significantly outperforms \exaqt{} across categories even as $k$ is increased for Hits@$k$.

\begin{figure}[t]
    \centering
    \includegraphics[scale=0.34]{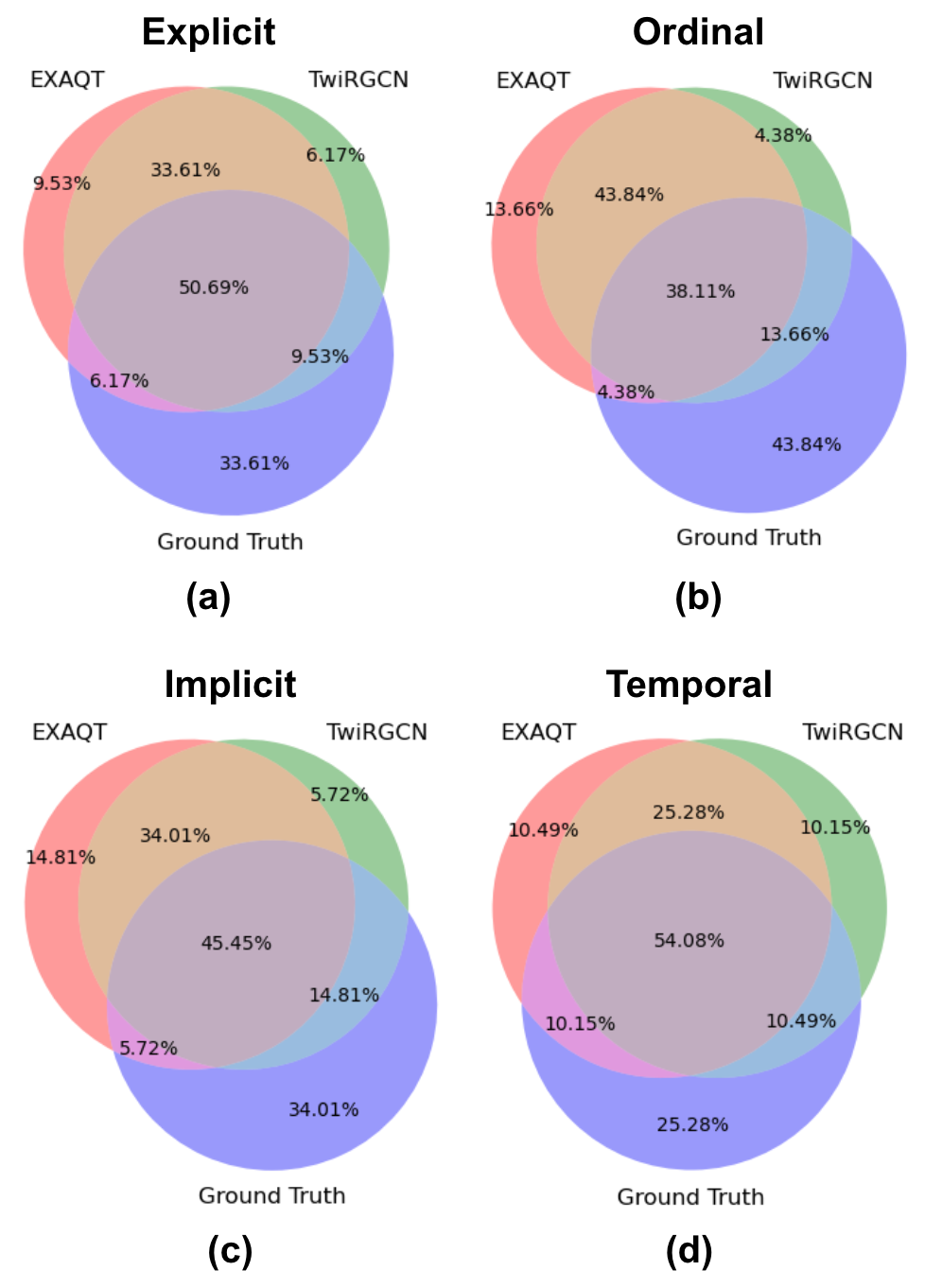}
    \caption{\label{fig:venn} \small Venn diagrams for the prediction overlap of EXAQT, ground truth and our best model for each category. 
    As described in \refsec{subsec:quali}, for Explicit, Implicit, and Ordinal question types \mdl{} gives the right answers for most questions that EXAQT answers correctly, while correctly answering a much larger set that EXAQT gets wrong.}
\end{figure}

 \noindent \textbf{Increasing temporal tolerance window:} In \reftbl{tbl:time_period}, we explore the effects of increasing the time window for marking an answer correct for temporal  questions. This means if the ground truth answer is 1992, and the predicted answer is 1990 for a question, it will be marked as incorrect in the $\pm1$ column and correct in the $\pm3$ column.  We find that our model, specifically \mdl{} (interval) gets even more accurate relative to \exaqt\ as we increase the temporal tolerance window.  This implies that \mdl{} is robust at ranking gold answers high up, even if they do not achieve rank~1.



\noindent \textbf{Prediction overlap:} We study the overlap of predictions between EXAQT, TwiRGCN, and ground truth. As seen in \reffig{fig:venn}, for Explicit, Implicit, and Ordinal question types our model gives the right answers for most questions that EXAQT answers correctly (missing less than $6\%$ on average), while correctly answering a much larger set that EXAQT gets wrong. This split is more even between the two models for the temporal-type questions.


\section{Conclusion}
\label{sec:End}

In this paper, we proposed \mdl, a TKGQA system that employs a novel, temporally weighted graph convolution for answering questions that require complex temporal reasoning over a TKG. \mdl{} modulates the convolutional messages through a TKG edge based on the relevance of the edge time interval to the question.  We present two temporal weighting schemes with complementary strengths, intuitively explained through their simple formulations.
We also propose an answer gating system for incorporating the pooled entity and time embeddings from \mdl{} in the prediction, based on the likelihood that the answer is a time or an entity, given the question.
Despite its relative simplicity, \mdl{} gives significantly superior TKGQA accuracy on a challenging dataset compared to more heavily engineered baselines.

\section*{Acknowledgements}

We thank Jeremy Cole and Srini Narayanan for their valuable feedback. We also thank the anonymous reviewers for their constructive comments. Soumen Chakrabarti was supported in part by grants from Amazon, Google, IBM, and SERB.



\section{Limitations}
\label{sec:Limits}


TwiRGCN is limited by the need for relevant subgraphs for each question to be provided in the dataset. Such subgraphs have been provided in the \data\ dataset used in the current work, but that may not be true for all TKGQA datasets.
This limitation may be addressed for datasets that do not provide subgraphs through recently proposed subgraph selection methods \citep{ref:science_dir_subgtr, shang-etal-2022-tsqa, exaqt}, but we leave that exploration for future work.

\bibliographystyle{styles/acl_natbib}
\bibliography{references.bib}

\clearpage

\appendix
\onecolumn
\begin{center}
\Large\bfseries\ztitle \\ \large (Appendix)
\end{center}

\section{Additional Analyses}
\label{appendix:analyses}

\subsection{Complete prediction overlap}
\label{appendix:overlap}
\begin{figure}
    \centering
    \includegraphics[scale=0.34]{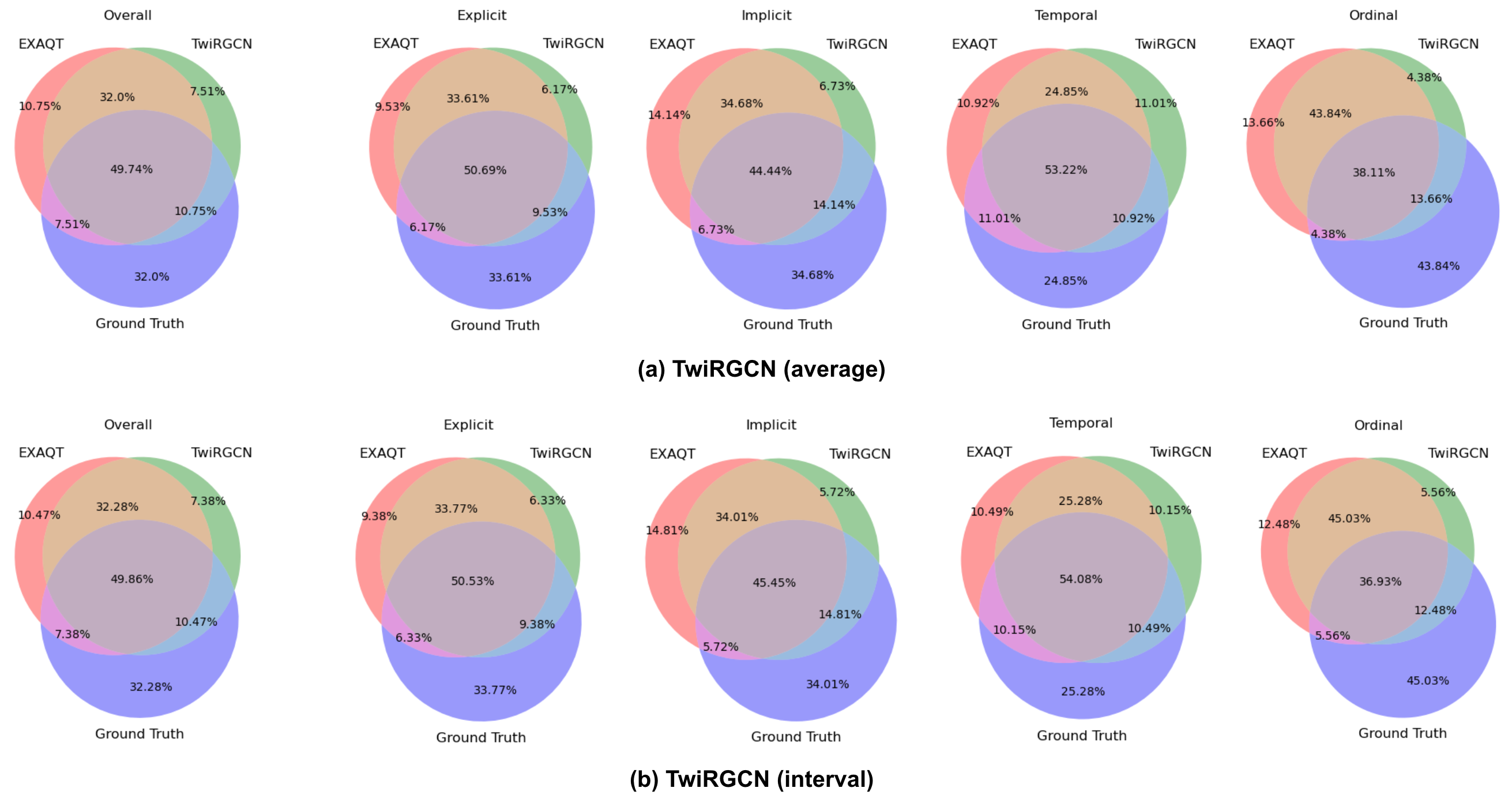}
    \caption{\label{fig:venn_full} Venn diagrams for the prediction overlap of EXAQT, ground truth, and our two models \mdl{} (average) in (a) and \mdl{} (interval) in (b), as discussed in \refappendix{appendix:overlap}.}
\end{figure}

In \reffig{fig:venn_full}, we extend our analysis in \ref{subsec:quali} by providing the complete prediction overlap for both our models with \exaqt{} and ground truth across all question categories in \data{}.

\section{Hyperprameters}
\label{appendix:hyperprameters}

\textbf{We use the following hyperparameters:}
\begin{itemize}[leftmargin=*]
\item Number of layers, $L = 2$
\item $c_d = 3$
\item train batch size = 32
\item valid batch size = 5
\item LR = 0.00004
\item Decay for LR = 0.4 every 10 epochs
\item Cosine distance scaling constant for training (described in \refsec{sec:method}) = 30
\end{itemize}
\par \smallskip
\noindent
\textbf{Model and program execution details:}
\begin{itemize}[leftmargin=*]
\item Number of parameters = 2,223,833
\item 11GB Nvidia GPU used with cudatoolkit 11.1
\item Time per training epoch = 1:04 min
\item Number of epochs to convergence on average = 50
\item Early stopping used and implemented in code with patience = 10
\item Validation overall Hits@1 for \mdl{} (average) = 0.606
\item Validation overall Hits@1 for \mdl{} (intervall) = 0.602
\item Performance is fairly stable around current hyperparameters without much tuning, except for LR decay rate. We used around 5--7 training runs with different decay settings to get the current rate. \mdl{} is stable around current settings.
\item Hyperparameters were tuned by manually inspecting loss behavior. Final values were selected based on a sustained, stable good performance on the test set for 3 runs.
\end{itemize}



\end{document}